\RequirePackage{fix-cm}
\documentclass[twocolumn]{svjour3}           
\smartqed  
\usepackage{natbib}
\usepackage{graphicx}
\usepackage{caption}
\usepackage{mathptmx} 
\usepackage{amsmath}
\usepackage{scrtime}
\usepackage[dont-mess-around]{fnpct}
\usepackage{etoolbox}
\usepackage{multirow} 
\usepackage{tabularx,ragged2e}

\newcolumntype{C}{>{\Centering\arraybackslash}X}  
\newcolumntype{b}{>{\Centering}X}
\newcolumntype{s}{>{\Centering\hsize=.5\hsize}X}
\newcolumntype{j}{>{\Centering\hsize=.4\hsize}X}
\newcolumntype{k}{>{\Centering\hsize=.3\hsize}X}
\newcolumntype{y}{>{\Centering\hsize=.2\hsize}X}
\newcommand*{\affaddr}[1]{#1}  
\newcommand*{\affmark}[1][*]{\textsuperscript{#1}}

\begin{document}

\title{Face Mask Extraction in Video Sequence}

\author{Yujiang Wang \protect\affmark[1] \and
        Bingnan Luo  \affmark[1] \and
        Jie Shen \affmark[1] \and
        Maja Pantic \affmark[1]
}

\institute{Yujiang Wang \at
              \email{yujiang.wang14@imperial.ac.uk}   \and    
           Bingnan Luo \at 
              \email{bingnan.luo16@imperial.ac.uk} \and
           Jie Shen \at  
            \email{jie.shen07@imperial.ac.uk}      \and
           Maja Pantic \at 
           \email{m.pantic@imperial.ac.uk}  \and
           \affaddr{\affmark[1]Department of Computing, Imperial College London, 180 Queens Gate, London, UK}
           }

\date{}

\maketitle

\begin{abstract}
Inspired by the recent development of deep network-based methods in semantic image segmentation, we introduce an end-to-end trainable model for face mask extraction in video sequence. Comparing to landmark-based sparse face shape representation, our method can produce the segmentation masks of individual facial components, which can better reflect their detailed shape variations. By integrating Convolutional LSTM (ConvLSTM) algorithm with Fully Convolutional Networks (FCN), our new ConvLSTM-FCN model works on a per-sequence basis and takes advantage of the temporal correlation in video clips. In addition, we also propose a novel loss function, called Segmentation Loss, to directly optimise the Intersection over Union (IoU) performances. In practice, to further increase segmentation accuracy, one primary model and two additional models were trained to focus on the face, eyes, and mouth regions, respectively. Our experiment shows the proposed method has achieved a 16.99\% relative improvement (from 54.50\% to 63.76\% mean IoU) over the baseline FCN model on the 300 Videos in the Wild (300VW) dataset.


\end{abstract}

\section{Introduction}
\label{intro}

The sparse facial shape descriptor extracted with traditional landmark-based face-tracker usually cannot capture the full details of the facial components' shapes, which are essential to the recognition of higher level features such as facial expressions, emotions, identity, and so on. To overcome the limitations of sparse facial descriptors, we introduce the concept of face mask, a dense facial descriptor with information of semantic facial regions at pixel level like eyes and mouth. Developing from various deep learning-based semantic image segmentation methods, we then propose a novel approach for extracting face mask in video sequence. Different from semantic face segmentation, face mask extraction handles occlusion in a similar way to facial landmark tracking. Namely, the extract face mask is expected to be complete regardless of occlusion, while typical segmentation result would exclude the occluded area. Face mask extraction techniques could have many potential and interesting applications in the field of Human-Computer Interaction, including face detection \& recognition, emotion \& expression recognition, social robots interaction, etc. To the best of our knowledge, this is the first exploration of face mask extraction in video sequence with an end-to-end trainable deep-learning model.  

Face mask extraction is a challenging task, especially for video clips taken in the wild, due to the huge amount of variations such as indoor \& outdoor conditions, occlusions, image qualities, expressions, poses, skin colours, etc. Early studies of semantic face segmentation \citep{kae2013augmenting,smith2013exemplar,lee2008markov,warrell2009labelfaces} usually concentrated on the segmentation of still face images, and their methods were mostly based on heavily engineered approaches rather than learning. 

In recent years, deep-learning techniques, particularly Convolutional Neural Networks (CNNs), has developed rapidly in the field of semantic image segmentation. Comparing to traditional approaches, the main advantage of deep-learning methods is their ability to learn robust representations through an end-to-end trainable model for a particular task and dataset, and their performances usually surpass that of hand-crafted features extracted by traditional computer vision method. Among others, Fully Convolutional Networks (FCN) \citep{long2015fully} is the first seminal work of applying deep-learning techniques in semantic image segmentation. FCN substitute the fully connected layers in the widely-used deep CNN architectures - such as AlexNet \citep{krizhevsky2012imagenet}, VGG-16 \citep{simonyan2014very}, GoogleLeNet \citep{szegedy2015going}, ResNet \citep{he2016deep} into convolutional layers, therefore turns the outputs from one-dimensional vectors to two-dimensional spatial heat-maps, which are then upsampled to the original image size using deconvolutional layers \citep{zeiler2011adaptive,zeiler2014visualizing}. Developed from the baseline FCN, many improvements have been proposed in the following years, achieving increasingly better performance on benchmark datasets. Some works have changed the decoder structure of FCN, like SegNet by \citet{badrinarayanan2017segnet}, and some other models have applied Conditional Random Field (CRF) as a post-processing step, such as the CRFasRNN work by \citet{zheng2015conditional} and the DeepLab models \citep{chen2016deeplab}, and there are also works that utilise dilated convolutions \citep{zhou2015exploiting}, or atrous convolutions in other words, to broaden the reception fields of filters without additional computation cost, e.g. the DeepLab models by \citet{chen2016deeplab}, ENet \citep{paszke2016enet} and the work of \citet{yu2015multi}.

Comparing to image segmentation, fewer works concern semantic segmentation in video sequences. Depending on the training methods, these works can be roughly divided into 1. fully-supervised methods \citep{kundu2016feature,liu2015multiclass,shelhamer2016clockwork,tran2016deep,tripathi2015semantic}, where all the annotations are given; 2. semi-Supervised approaches \citep{jain2014supervoxel,nagaraja2015video,tsai2016video,caelles2017one}, which require certain pixel-level annotations like the ground truth of the sequence's first frame; and 3. weakly-supervised ones \citep{saleh2017bringing,drayer2016object,liu2014weakly,wang2016semi}, in which only the tags for each video clips are known. Due to the complex variations in real-life scenarios, we focus on fully-supervised video semantic segmentation. In addition, most semi-supervised or weakly-supervised approaches are proposed to solve the task of video object segmentation, i.e. binary classification between foreground and background, which limits their application in multi-class tasks such as face mask extraction.

To utilise the temporal information in video sequences, several fully-supervised video segmentation methods rely on graphical models such as \citet{kundu2016feature,liu2015multiclass,tripathi2015semantic}, while other approaches are based on CNN models, e.g. the Clockworks Convnets by \citet{shelhamer2016clockwork}, in which a fixed or adaptive clock was used to control the update rates of different layers according to their semantic stability. Other works, such as
\citet{zhang2014discriminative} and \citet{tran2016deep}, use 3D convolutions or 3DCNNs to capture the temporal dependencies as well as the spatial connections. Both approaches have their limitations. Clockworks Convnets do not fully utilise the temporal information in video sequence since the semantic changes are only used to adjust clock rates. 3DCNN treats temporal dimension in the same way as 2D space, thus could limit the extraction of long-term temporal information. 

In this paper, we propose an end-to-end trainable model which could exploit the temporal information in a more direct and natural way. The key idea is the application of Convolutional Long Short Term Memory (ConvLSTM) layer \citep{xingjian2015convolutional} in FCN models, which enable the FCNs to learn the temporal connections while retaining the ability to learn spatial correlations. 

Recurrent Neural Networks, especially LSTMs, have already shown their capabilities to capture short and long term temporal dependencies in various computer vision tasks such as visual speech recognition \citep{lee2016multi,zimmermann2016visual,chung2016out,petridis2017end,petridis2017end2}. However, typical RNN models only accept one-dimensional arrays, which limits the models' application in tasks that require multi-dimensional relationships to be kept. To overcome this limitation, multiple approaches have been proposed, such as the works of \citet{graves2007multi}, the ReNet architecture of \citet{visin2015renet}, and the aforementioned ConvLSTM by \citet{xingjian2015convolutional}. 

Among these methods, ConvLSTM directly models the spatial relationships while keeping LSTM's ability to capture temporal dependencies. Another advantage of ConvLSTM is it can be integrated into existing convolutional networks with very little effort because a convolutional layer can be easily replaced by a ConvLSTM layer with identical filter settings. 

In this work, we introduce the ConvLSTM-FCN model that combines FCN and ConvLSTM by converting a certain convolutional layer in the FCN model into a ConvLSTM layer, thus adding the ability to model temporal dependencies within the input video sequence. Specifically, for the baseline model, we adopt the structure of FCN model based on ResNet-50 \citep{he2016deep} and then replace the classifying convolutional layer, which is converted from the fully connected layer in the original ResNet-50 model, with a ConvLSTM layer with the same convolutional filter settings. We also add two reshape layers since ConvLSTM layers require different input dimensions than the convolutional layers. The ConvLSTM-FCN model accepts video sequence as input and outputs the predictions of the same size, and the temporal information is learnt together with the spatial connections. 

To be able to optimise the model toward higher accuracy in terms of mean Intersection over Union (mIoU), which is a typical performance metric for segmentation problems, we also propose a new loss function, called Segmentation Loss. Unlike the IoU loss in \citet{rahman2016optimizing}, Segmentation Loss is more flexible and carries more practical meaning in image space. In comparison to the frequently-used cross-entropy loss and the IoU loss by \citet{rahman2016optimizing}, higher mIOU can be achieved when Segmentation Loss is used as the loss function during training.

A dataset with fully annotated face masks in videos would be needed to evaluate the proposed method. However, at this moment, no such dataset could be found in the public domain. Therefore, in this work, we use the 300 Videos in the Wild (300VW) dataset \citep{shen2015first}, which contains per-frame annotations of 68 facial landmarks for 114 short video clips. These landmark annotations are then converted into 4 semantic facial regions: face skin, eyes, outer mouth (lips) and inner mouth.

Our experiments are conducted on the aforementioned 300VW dataset with converted pixel-level labels of 5 class (the 4 facial regions plus background). As the baseline approaches, we compare performances of 1. The traditional 68-point facial landmark tracking model \citep{kazemi2014one}; 2. The deeplab-V2 model \citep{chen2016deeplab}; 3. The VGG-16 Version of FCN \citep{simonyan2014very,long2015fully},  4. The ResNet-50 Version FCN \citep{he2016deep,long2015fully}, and 5. The ResNet-50 Version FCN + a simple temporal smoothing strategy. We then change the ResNet-50 version FCN to ConvLSTM-FCN, so that the temporal information in video sequence could be utilised. For better performance, we further extend our method to include three ConvLSTM-FCN models: a primary model to find the face region, and two additional models focusing on the eyes and mouth, respectively. The predictions of the three models are combined to obtain the final face mask. Our experimental results show that the utilisation of temporal information could significantly improve FCN's performances for face mask extraction (from 54.50\% to 63.76\% mean IoU), and the performance of ConvLSTM-FCN model also surpass that of traditional landmark tracking models (63.76\% Versus 60.09\%). 

\section{Related Works}
\label{sec:RW}
This section covers the major related works in the field. It is worth mentioning that, to the best of our knowledge, there is no similar work in terms of semantic face segmentation or face mask extraction in video sequence, so we have investigated the studies of video semantic segmentation instead.

\subsection{Semantic Image Segmentation}
\label{sec:SIS}
The last few years have witnessed the rapid development of deep-learning techniques in the field of semantic image segmentation, and most of the state-of-the-art results are achieved by such models. The FCN by \citet{long2015fully} is the first milestone for deep learning in this field. FCN cast the fully convolutional layers in well-known deep architectures, such as AlexNet \citep{krizhevsky2012imagenet}, VGG-16 \citep{simonyan2014very}, GoogleLeNet \citep{szegedy2015going}, ResNet \citep{he2016deep}, to convolutional layers so that the output of such models is spatial heat-maps instead of traditional one-dimensional class score. The skip-architecture of FCN enables the information from coarser layers to be seen by finer layers, therefore the model can be more aware of the global context, which is rather important in semantic segmentation. FCNs have limitations in term of integrating knowledge of the global context to make appropriate local predictions since the receptive field of their filters can only increase linearly when the number of layers grows \citep{garcia2017review}. Therefore, later studies improve their models' abilities to utilise the global image context with different approaches. 

The works of the DeepLab models \citep{chen2016deeplab}, ENet \citep{paszke2016enet} and the work of \citet{yu2015multi} has involved the application of dilated convolutions, or so-called atrous convolutions. They are a kind of generalised Kronecker-factored convolutional filters \citep{zhou2015exploiting}, and they differ from traditional convolutional filters in that they have wider receptive fields which can grow exponentially with the dilated rate \emph{l} \citep{garcia2017review}. The standard convolutional operations can be seen as dilated convolutions with dilated rate = 1. Dilated convolutional layers can have more awareness of the global image context without reducing the resolution of feature maps too much. Another noticeable improvement is brought by the works of \citet{yu2015multi}, where their models take inputs of images at two different scales and then combine the predictions into one. The ideas of integrating predictions from multi-scale images can also be seen in the works of \citet{roy2016multi} and \citet{bian2016multiscale}. 

Conditional Random Field (CRF) is a frequently-used technique for deep semantic segmentation models, such as the DeepLab models \citep{chen2016deeplab} and the CRFasRNN by Zheng et al. \citet{zheng2015conditional}. The main advantage of CRF is that it could capture the long-range spatial relationships which are usually difficult for CNNs to retain, and CRF could also help to smooth the edges of the predictions. 

\subsection{Semantic Face Segmentation}
\label{sec:SFS}
Most earlier works of semantic face segmentation applied engineering-based approaches. \citet{kae2013augmenting} employed a restricted Boltzmann machine to build the global-local dependencies such that the global shape can be natural, while they used CRFs to construct the details of the local shape. As in the work of \citet{smith2013exemplar}, a database of exemplary face images was first collected and labelled, and face images were aligned to those exemplary images with a non-rigid warping. There are also some other earlier works \citep{warrell2009labelfaces,scheffler2011joint,yacoob2006detection,lee2008markov} in this field, however, most such works utilised engineering-based hand-crafted features, and it usually takes lots of time to fine-tune those models for them to work under particular scenarios. Therefore, they were gradually replaced by deep-learning based approaches.

Compared with the rapid progress of deep learning in semantic image segmentation, its application in semantic face segmentation is comparatively rare. Due to the difficulties of pixel-level labelling for huge amounts of data, currently, there are only a few publicly available datasets for this task. Two commonly used datasets are Parts Label dataset \citep{learned2016labeled,kae2013augmenting}, which contains 2927 images with labels of background, face skin and hair, and Helen dataset \citep{le2012interactive,smith2013exemplar} including 2330 face images with annotations of face skin, left/right eyebrow, left/right eye, nose, upper lip, inner mouth, lower lip and hair. The lack of public face datasets with pixel-level annotations could be an obstacle for the development of deep models in this field.

For those face segmentation approaches using deep models, the works of \citet{zhou2015interlinked} proposed an interlinked version of the traditional CNN model, where parts of the face could be detected except the facial skin. Compared with FCN, the proposed model is less efficient and its structure is overly redundant, and it cannot detect semantic part at large scales, like the facial skin. \citet{gucclu2017end} took advantages of multiple deep-learning techniques, i.e. they formulated a CRF by one Convolutional Neural Network for the unary potential and the pairwise kernels, and one Recurrent Neural Networks to transform the unary potentials and the pairwise kernels into segmentation space. The training process utilised the idea of Generative Adversarial Networks (GAN), where the CRF and a discriminator network played a two-player minimised game. The limitation of this work is that it requires an initial face segmentation generated by a facial landmark detection model as the input in addition to the original face image, while the initial face segmentation is not necessary in our method.

All these semantic face segmentation approaches were proposed for still face images, while in the context of video sequences, where the variations are more complex, these methods may not be applicable. Currently, to the best of our knowledge, our work is the first one developed for semantic face segmentation in video sequence, or face mask extraction as we propose. 

\subsection{Video Semantic Segmentation}
\label{sec:VSS}
Video semantic segmentation methods can be roughly separated into three types through their supervision settings, which are: 1. The works that handle fully-supervised problems, i.e. the pixel-level annotations of all frames are known, 2. The semi-supervised video segmentation approaches, in which partial pixel-level annotations are known, such as only the ground-truths of the first frame is known for both training and testing, 3. The weakly-supervised methods focus on scenarios where only the tags of each video are given for the learning process. The main-stream interest of video segmentation community is on the semi-supervised problems \citep{jain2014supervoxel,nagaraja2015video,tsai2016video,caelles2017one} and the weakly-supervised issues \citep{saleh2017bringing,drayer2016object,liu2014weakly,wang2016semi}, while the tasks of these problems are usually about segmenting one single object out of the background in a video sequence. This is somehow different from the scenarios of face mask extraction, where multiple semantic face parts should be extracted. Therefore, we have investigated the less-focused fully-supervised video segmentation works.

Some of these fully-supervised works replied on graphic models \citet{kundu2016feature,liu2015multiclass,tripathi2015semantic}. As for these approaches using deep models, the idea Clockworks Convnets by \citet{shelhamer2016clockwork} was based on the observation that the semantic contents of two successive frames change relatively slower than pixels. The proposed Clockworks Convnets used a clock at either fixed or adaptive schedules to control the update rates of different layers basing on the semantic content evolution. This work does not fully utilise the temporal information. The works of \citet{zhang2014discriminative} and \citet{tran2016deep} have both shown the idea of applying 3DCNN or 3D convolutions to capture information at time dimension. Treating temporal dependencies in the same way as spatial connections may hinder the model to understand some subtle temporal information, and they may not be able to capture the long-term time dependencies.

In our model, the temporal dependencies are extracted in a more natural and effective approach, through the application of Convolutional LSTM. 

\subsection{Convolutional LSTM}
\label{sec:CLSTM}
Convolutional LSTM (ConvLSTM) is proposed by \citet{xingjian2015convolutional} to solve the problem of precipitation nowcasting. Its has a similar structure as the FC-LSTM by \citet{graves2013generating}, while all the inputs $X_1$, \ldots, $X_t$, cell outputs $C_1$, \ldots, $C_t$, hidden states $H_1$, \ldots, $H_t$, input gate $i_t$, forget gate $f_t$ and output gate $o_t$ in ConvLSTM are 3D tensors, where the first dimension is the measurements in cell varying over time, and the last two dimension are spatial ones (rows and columns) \citep{xingjian2015convolutional}. The key idea of ConvLSTM can be expressed in Eq. \ref{eq:1} \citep{xingjian2015convolutional}, where '$\ast$' denotes the convolutional operator and '$\circ$' means the Hadamard product.

\begin{align}
\begin{split}
\label{eq:1}
 i_t &= \sigma(W_{xi}\ast X_t+W_{hi}\ast H_{t-1}+W_{ci}\circ C_{t-1}+b_i) \\
 f_t &= \sigma(W_{xf}\ast X_t+W_{hf}\ast H_{t-1}+W_{cf}\circ C_{t-1}+b_f) \\
 C_t &= f_t \circ C_{t-1} + i_{t} \circ tanh(W_{xc}\ast X_t+W_{hc}\ast H_{t-1}+b_c) \\
 o_t &= \sigma(W_{xo}\ast x_t+W_{ho}\ast H_{t-1}+W_{co}\circ C_t+b_o) \\
 H_t &= o_t \circ tanh(C_t)  \\
\end{split}
\end{align}

ConvLSTM could capture the long and short term temporal dependencies while retaining the spatial relationships in the feature maps, therefore it is an ideal candidate for face mask extraction in video sequence. Besides, with these convolutional operations in cells, a standard convolutional layer could be easily cast into a ConvLSTM layer with identical convolutional filters. Due to these advantages, we have utilised ConvLSTM in FCN structures to understand the temporal dependencies in video sequence.

\subsection{Cascade models for coarse-to-fine predictions}
\label{sec:CAS_Model}

The ideas of using cascade deep models to gain coarse-to-fine predictions have been used widely by various works \citep{sun2013deep,zhang2014coarse,zhou2013extensive,zhang2016joint} on facial landmark localization and face alignment.  For instance, the work of \citet{sun2013deep} adapted a three-level cascade CNN models to detect facial landmarks. In this work, a first-level Convolutional Network was trained to locate global key-points over the whole faces, and the local areas around these predictions were input into the CNNs of the next two levels to obtain landmarks with better qualities. Similar ideas was employed in the work of \citet{zhou2013extensive} where a four-level regressive CNN model was demonstrated for extensive facial landmark localisation. The initial landmark predictions with less accuracy were made by the second-level model (first-level was for bounding-box detection), and the facial components were cropped using those predictions and were input into later models to refine the landmark qualities. \citet{zhang2014coarse} proposed a cascade Coarse-to-Fine Auto-Encoder Network for the tasks of face alignment. The first Auto-Encoder model generated global landmarks with lower quality, and these key-points are gradually refined by the following Auto-Encoders which zoomed in the local regions around the last model's predictions as their inputs. 

To gain better performances, we draw from these works the ideas of using cascade models for coarse-to-fine predictions and apply it in our tasks. Particularly, we have employed an engineering trick of utilising a primary model for whole-face predictions and then training two zoomed-in models to refine local predictions on eye and mouth regions, respectively.

\section{Methodology}
\label{sec:MT}
The section explains our proposed ConvLSTM-FCN model and the segmentation loss function. In addition, we also introduce the engineering trick of combining the additional eye and mouth models with the primary model.

\subsection{ConvLSTM-FCN Model}
\label{sec:MT_Model} 

The first FCN model based on VGG-16 \citep{long2015fully} was proposed in 2015. Many variations of the FCN model have been developed afterward, usually achieving higher performances and better training efficiency. 

In this work, we base our model on the structure of the FCN model released by \citet{Keras-FCN}. This model is a ResNet-50 version FCN. The details about this model's structure are summarised in Table \ref{tab:1}. Compared with the standard ResNet-50 architecture \citep{he2016deep}, dilated convolutions with dilated rate = 2 are used in the building blocks of 'Conv5\_x' layer instead of the ordinary convolutional operations. The 'Conv6' layer is the classifying layer which replaces the original fully-connected layer to produce feature maps of size 20$\times$20 at C channels, where C is the number of target classes. A bi-linear up-sampling layer of 16s is used instead of a deconvolutional layer. 

\begin{table}[hb!]
\caption{Architectures of the baseline FCN model. This model adopts the input size of 320*320. Building blocks are illustrated in brackets with the number of stacked blocks. The structures of building blocks at 'Conv1', 'Conv2\_x', 'Conv3\_x' and 'Conv4\_x' layers are identical to the original ResNet-50 model, while in 'Conv5\_x' layer, atrous convolutional filters with dilated rate = 2 are used instead of the standard convolutions. The 'Conv6' layer is the classifying layer that outputs feature maps at C channels, where C is the number of target classes. The 'UpSampling' layer bi-linearly up-samples the feature maps back to the input size at 16s up-sampling rate.}
\label{tab:1} 
\begin{tabularx}{\columnwidth}{ l | b | s | s}
\hline
Layer Name & Building Blocks & Output Size & Dilated Rate \\ \hline
Conv1 & 7$\times$7, 64, stride 2
& 160$\times$160 & 1$\times$1 \\ \hline
\multirow{2}{*}{Conv2\_x} & \multicolumn{3}{c}{3$\times$3 max pooling, stride 2} \\ \cline{2-4}
 & $\left[\begin{array}{cc} 1\times 1, & 64  \\3\times 3, & 64 \\1\times 1, & 256 \end{array}\right]\times 3$  & 79$\times$79 & 1$\times$1   \\ \hline 
Conv3\_x & 
$\left[\begin{array}{cc} 1\times 1, & 128  \\3\times 3, & 128 \\1\times 1, & 512 \end{array}\right]\times 4$  
& 40$\times$40 & 1$\times$1 \\ \hline
Conv4\_x & 
$\left[\begin{array}{cc} 1\times 1, & 256  \\3\times 3, & 256 \\1\times 1, & 1024 \end{array}\right]\times 6$  
& 20$\times$20 & 1$\times$1 \\ \hline
Conv5\_x & 
$\left[\begin{array}{cc} 1\times 1, & 512  \\3\times 3, & 512 \\1\times 1, & 2048 \end{array}\right]\times 3$  
& 20$\times$20 & 2$\times$2 \\ \hline
Conv6 & 1$\times$1, C, stride 1
& 20$\times$20 & 1$\times$1 \\ \hline
UpSampling & None
& 320$\times$320 & None \\ \hline
\end{tabularx}
\end{table}

\begin{figure}[ht!]
 \includegraphics[width=\columnwidth]{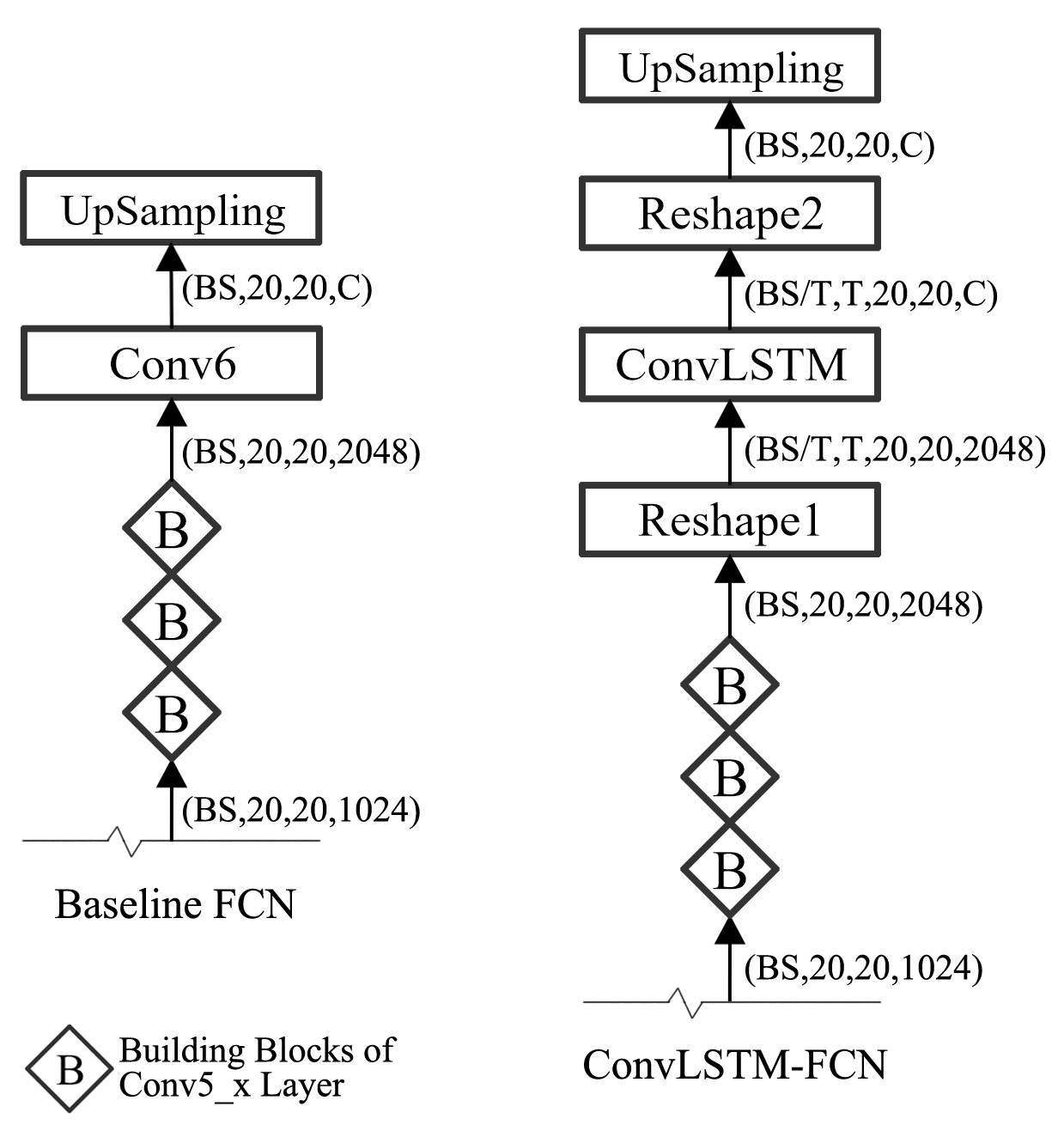}
\caption{An illustration of casting baseline FCN into ConvLSTM-FCN model. Only these top layers are shown. 'BS' refers to batch size of images, 'T' denotes the time dimension in ConvLSTM layer and 'C' is the target classes number. The ConvLSTM layer in ConvLSTM-FCN has the same convolutional filters with Conv6 layer in Baseline FCN. Two reshape layers are added to convert tensor dimensions in ConvLSTM-FCN.}
\label{fig:1}       
\end{figure}

The conversion of baseline FCN to ConvLSTM-FCN is performed by replacing the 'Conv6' layer with a ConvLSTM layer of identical convolutional filters. Fig. \ref{fig:1} shows the details of this procedure. The Reshape1 layer is used to output tensor with one additional time dimension 'T', which is required by the ConvLSTM layer, and the Reshape2 layer cast the tensor back. 'T', the time dimension in the ConvLSTM layer, refers to the number of frames in a video sequence. Therefore, for the ConvLSTM-FCN model to work effectively, the image orders within one batch should be arranged properly so that ConvLSTM layer could accept video sequences in the correct format.

\subsection{Segmentation Loss}
\label{sec:MT_Loss}
This section introduces the new loss function that we propose to optimise mean Intersection over Union (mIoU).

MIoU is the most frequently-used performance metric in the field of semantic segmentation. For one annotation set and its predictions, IoU is calculated by the intersection divided by the union. The intersection is actually the true positives of the confusion matrix, while the union is the sum of true positives, false positives and false negatives. mIoU is the average of IoUs over all non-background classes. Assuming there are a total of C non-background classes, and the notation $n_{ij}$ stands for the number of pixels whose annotation is $i$ with prediction $j$, then mIoU can be expressed in Eq. \ref{eq:2}.

\begin{align}
\label{eq:2}
\displaystyle
mIoU = \displaystyle\frac{1}{C} \displaystyle\sum_{i=1}^{C}{\displaystyle\frac{n_{ii}}{\displaystyle\sum_{j=1}^{C}{n_{ij}} + \displaystyle\sum_{j=1}^{C}{n_{ji}} - n_{ii} }}
\end{align}

The main reason for using mIoU as the metric of segmentation accuracy instead of Classification Rate (CR) is to avoid the bias caused by class imbalances. Class imbalance is a common and challenging problem in semantic segmentation. For example, a face image usually contains much fewer eye pixels than background pixels. If all eye and background pixels are predicted as background, the resulting CR will still be quite high, which is unfair and misleading. In contrast, mIoU would be 0 in such case as there would be no true-positive for the eye pixels. Therefore, in the field of semantic segmentation, mIoU is used as the main evaluation metric, and its performance is not directly related to CR.

Cross-entropy loss, or softmax loss, is one of the most widely-used loss function in deep learning. Although cross-entropy loss is a useful loss with smooth training curves, it drives the model toward higher average Classification Rate (CR), which does not necessarily lead to improvement in mIoU. In other words, using cross-entropy loss in semantic segmentation could not fully fulfil deep models' potential in the task. Therefore, we propose a new loss, which we name as Segmentation Loss, to optimise the model's mIoU performances directly.

The work of \citet{rahman2016optimizing} has used a similar idea of optimising IoU using an IoU Loss instead of cross-entropy loss. One immediate limitation of the IoU Loss is that it can only be applied to binary segmentation tasks, i.e. the background / foreground segmentation problems. Extending the loss formulation to multiple-class scenarios is straightforward. In particular, let $PR_i^t$ be the model’s prediction (output of Softmax) for the $i^{th}$ sample belong to class $t$, and denote $GT_i^t$ as the binary class annotation for the $i^{th}$ sample to be class $t$ (i.e. 1 if the sample actually belongs to class $t$ and vice versa), and there are a total of $C$ classes and $K$ samples, the multiple-class IoU Loss could be expressed in Eq. \ref{eq:9}. 

\begin{align}
\begin{split}
\label{eq:9}
\displaystyle
IoULoss_{mutiple} &= 1-\frac{1}{C}\displaystyle\sum_{t=1}^{C}{\frac{\displaystyle\sum_{i=1}^{K}{(PR_i^t*GT_i^t)}}{\displaystyle\sum_{i=1}^{K}{(PR_i^t+GT_i^t-PR_i^t*GT_i^t)}}}
\end{split}
\end{align}

The multiple-class IoU Loss in Eq. \ref{eq:9} is essentially a kind of 'soft' mean IoU with computable derivatives, and it is a natural extension from the binary IoU Loss of \citet{rahman2016optimizing}. This multiple-class IoU Loss is one of the baselines for comparing the performances of our Segmentation Loss. 

Another drawback of the IoU Loss proposed in \citet{rahman2016optimizing} is that it neglects the practical meaning of the IoU gradient, and, as a result, takes an over-simplified form. This is shown in the following analysis.

Consider the case of single class segmentation, where annotations is either 1 (foreground, positive samples) or 0 (background, negative samples). Denote predictions as A, ground-truths as B and the network parameters as $\theta$. Let $g(\theta)=A\cap B$ and $f(\theta)=A\cup B$, then this single-class IoU can be expressed as in Eq. \ref{eq:3}:

\begin{align}
\begin{split}
\label{eq:3}
IoU = \frac{A\cap B}{A\cup B}=\frac{g(\theta)}{f(\theta)} \\
\end{split}
\end{align}

If we treat IoU as the direct objective function, we need to find IoU's gradient, which is denoted as $(IoU)'$, in order to optimise this objective function. The deduction of $(IoU)'$ is shown in Eq. \ref{eq:4}.

\begin{align}
\begin{split}
\label{eq:4}
\displaystyle
(IoU)' = (\frac{g(\theta)}{f(\theta)})'=\frac{f(\theta)g'(\theta)-g(\theta)f'(\theta)}{f^2(\theta)} \\
\noindent=\frac{1}{f(\theta)}g'(\theta)+\frac{g(\theta)}{f^2(\theta)}(-f'(\theta))  \\
\end{split}
\end{align}

The work of \citet{rahman2016optimizing} set the value of $g'(\theta)$ to 0 for pixels where ground-truths is 0, while $f'(\theta)$ is set to 0 for positive samples. However, we argue that the $g'(\theta)$ and $f'(\theta)$, which is the gradient for $g(\theta)$ and $f(\theta)$, hold their practical meanings in IoU optimisation and should not be simplified in this approach. 

Since $g(\theta)=A\cap B$, for the purpose of optimising IoU, an appropriate gradient $g'(\theta)$ should encourage the predictions of the positive samples to  change from 0 to 1. Similarly, for $f(\theta)=A\cup B$, the gradient ($-f'(\theta)$) should drive the prediction of negative samples' from 1 to 0. From this perspective, $g'(\theta)$ stands for the optimisation direction of positive samples, while ($-f'(\theta)$) reveals how to optimise negative samples. With these discoveries, we could reformulate the loss function regarding IoU in a meaningful and natural way. Assuming there are a total of $K$ samples and ${x_i}$ is the $i^{th}$ sample, if we let $W_p=\frac{1}{f(\theta)}$ and $W_n=\frac{g(\theta)}{f^2(\theta)}$, the proposed Segmentation Loss function can be found in Eq. \ref{eq:5}.

\begin{align}
\begin{split}
\label{eq:5}
\displaystyle
SegLoss = \sum_{i=1}^{K}{W_p I_1(x_i)L_p(x_i)} + \sum_{i=1}^{K}{W_n I_0(x_i)L_n(x_i)}  
\end{split}
\end{align}

In Eq. \ref{eq:5}, $I_1(x_i)$ and $I_0(x_i)$ are the indicator functions for positive and negative samples respectively, and $L_p(x_i)$ and $L_n(x_i)$ are certain types of loss calculation functions for positive and negative samples separately. 

Extending Eq. \ref{eq:5} to the case of total C classes and performing the normalisation, we can express the complete form of Segmentation Loss in Eq. \ref{eq:6}. $I_1^t(x_i^t)$ is now the indicator function for the positive samples of class $t$, and vice versa for $I_0^t(x_i^t)$. 

\begin{align}
\begin{split}
\label{eq:6}
\displaystyle
SegLoss = \frac{\displaystyle\sum_{t=1}^{C}{\displaystyle\sum_{i=1}^{K}({W_p^t I_1^t(x_i^t)L_p(x_i^t)} + W_n^t I_0^t(x_i^t)L_n(x_i^t)})}{K\displaystyle\sum_{t=1}^{C}{(W_p^t+W_n^t})}
\end{split}
\end{align}

It can be seen from Eq. \ref{eq:6} that, in our Segmentation Loss, the loss of positive and negative samples from different classes is weighted separately by $W_p^t$ and $W_n^t$, and these weights are somehow related to the number of samples over different classes. For example, if there are fewer samples belonging to class $t$, its positive samples are more likely to hold a larger weight $W_p^t$, since the union of class $t$ can be smaller than that of other classes. Therefore, our Segmentation Loss has properly considered the imbalanced data distributions over different classes, which are ignored in cross-entropy loss. Also, the Segmentation Loss is a more comprehensive loss definition for IoU optimisation when compared with the work of \citet{rahman2016optimizing}.

The loss calculation function for positive and negative samples, which is $L_p(x_i)$ and $L_n(x_i)$ in Eq. \ref{eq:5} and Eq. \ref{eq:6}, could have a variety of potential definitions. In this paper, we have provided two different definitions for them. Their first definition, which can be seen as a variant form of categorical hinge loss, is shown in Eq. \ref{eq:7}.

\begin{align}
\begin{split}
\label{eq:7}
\displaystyle
L_p(x_i) &= max(max(PR_i\circ (GT_i)^{-1})-PR_i\cdot GT_i+g),0)  \\
L_n(x_i) &= max(PR_i\cdot oneHot(t)-PR_i\cdot GT_i+g,0) \\
\end{split}
\end{align}

In Eq. \ref{eq:7}, $GT_i$ and $PR_i$ are both $1\times C$ vectors, where $PR_i$ is the model's prediction for the $i^{th}$ sample $x_i$, e.g. (-1.2,2.9,7.1) for a 3-class sample, and $GT_i$ is the sample's ground truth as a one-hot vector, such as (0,1,0) for a ground truth of 2 with total 3 classes. $(GT_i)^{-1}$ refers to the inverse of $GT_i$, for example, if $GT_i=(0,1,0)$, then $(GT_i)^{-1}=(1,0,1)$. $oneHot(t)$ casts the number $t$ into the one-hot vector, and $max(a,b,c,\ldots)$ returns the maximum element. $g$ is a positive constant used to increase the discriminativities of loss function. The symbol '$\circ$' represents vector's Hadamard (element-wise) product, while '$\cdot$' means the dot product.

A second definition of $L_p(x_i)$ and $L_n(x_i)$ can be found in Eq. \ref{eq:8}, where the meanings of $PR_i$, $GT_i$, $g$ and $oneHot(t)$ remain unchanged. The intuitions of this definition are straight-forward, encouraging the predicted values of ground truth class to increase and penalising for those false negative classifications.

\begin{align}
\begin{split}
\label{eq:8}
\displaystyle
L_p(x_i) &= -(PR_i\cdot GT_i)  \\
L_n(x_i) &= 
    \begin{cases}
      0, & \text{if}\ PR_i\cdot GT_i> PR_i\cdot oneHot(t) + g\\
      PR_i\cdot oneHot(t) & \text{otherwise}
    \end{cases} \\
\end{split}
\end{align}

\subsection{Primary and Zoomed-in Models}
\label{sec:MT_PAM}

\begin{figure}[ht!]
\includegraphics[width=\columnwidth]{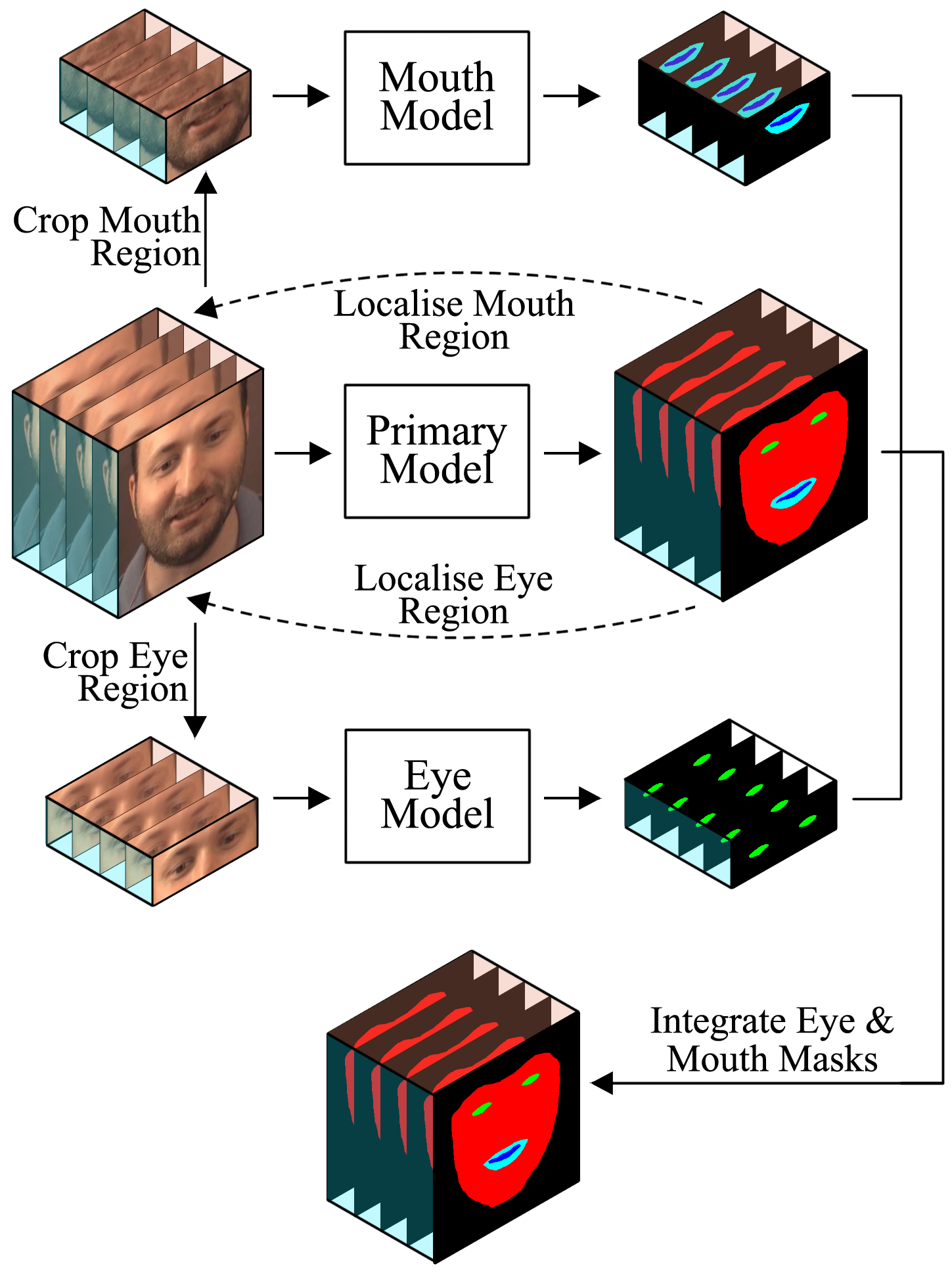}
\caption{An illustration of how the primary and zoomed-in models work. The primary face masks are extracted out of the video sequence by the primary model, and these masks are used to localise the mouth and eye regions. The cropped mouth and eye sequences are then fed into the additional mouth and eye models, respectively, to extract mouth and eye masks at higher accuracies. The primary face mask, eye and mouth masks are then combined to obtain the final face mask. (Best seen in colour)}
\label{fig:2}       
\end{figure}

\begin{figure*}[ht!]
\captionsetup{justification=centering}
\includegraphics[width=\textwidth]{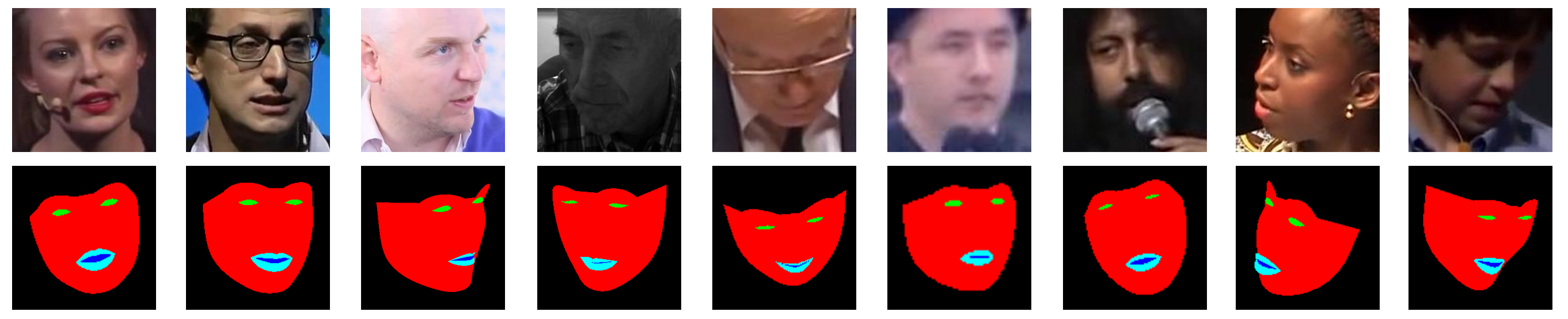}
\caption{Several examples of face images/masks from the 300VW dataset. Each column is a pair of face image/mask. The colours red/green/cyan/blue in face masks stands for facial skin/eye/outer mouth/inner mouth, respectively. (Best seen in colour)}
\label{fig:3}       
\end{figure*}

In practice, to further increase segmentation accuracy, we have trained one primary model for initial face mask extraction and two additional models to focus on the eyes and mouth region, respectively. Particularly, the primary model takes a face video sequence and outputs face masks for each frame, and these face masks are used to localise and crop the eye and mouth regions out of the video sequence. Two additional trained models, one for eye and another for mouth region, are then used to generate the eye and mouth masks, which are usually more accurate than the corresponding regions in the primary face mask. The final predictions are obtained from the outputs of the three models, i.e. the eye and mouth masks are mapped back to the primary face mask, replacing these corresponding areas. The pipeline of how primary and additional models work is shown in Fig. \ref{fig:2}.

\section{Experiments}
\label{sec:EXP}

\subsection{Dataset}
\label{sec:Dataset}

All our experiments are implemented on the 300 Videos in the Wild (300VW) dataset \citep{shen2015first}. The 300VW dataset consists of 114 videos taken in unconstrained environments and the average duration of each video clip is 64 seconds with a frame rate of 30 fps. All 218595 frames in these videos have been annotated manually with the 68 facial landmarks as in the works of \citet{sagonas2013300,sagonas2013semi}. The scenarios of this dataset can be roughly divided into three categories with increasing challenges: 1. Category one where videos are taken under conditions with good lightings and potential occlusions such as glasses or beard may occur. 2. Videos of category two can have larger variations than category one, e.g. in-door environment without enough illumination, overly-exposed cameras, etc. while the occlusions are similar. 3. Category three is the most challenging one, with videos of high variations from totally unconstrained environments. 

In order to obtain the face mask ground truths of all frames in the 300VW dataset, we have converted the 68-landmark annotations into pixel-level labels of one background class and four foreground classes: facial skin, eyes, outer mouth and inner mouth. This is achieved using cubic spline interpolation (with relaxed continuity constraints on eye corners and mouth corners) on corresponding landmark points. The generated face masks do not contain the nose region, since the the 68-landmark annotations do not cover the full boundary of noses. Besides, the nose is not an essential facial component for the face mask extraction techniques to be applied in Face Expression Recognition (FER), since it carries far less emotional information than other facial regions like eyes and mouths, and the benefits brought by annotating nose masks are comparatively small. Therefore, considering the costs and the benefits, no nose regions are included in our generated face masks. 

Some examples of the obtained face masks are shown in Fig. \ref{fig:3}. It can be seen from the figure that some videos in 300VW are quite challenging due to the high variations in head pose, illumination, occlusion, video resolution, etc. 

\begin{figure}[!hb]
\includegraphics[width=\columnwidth]{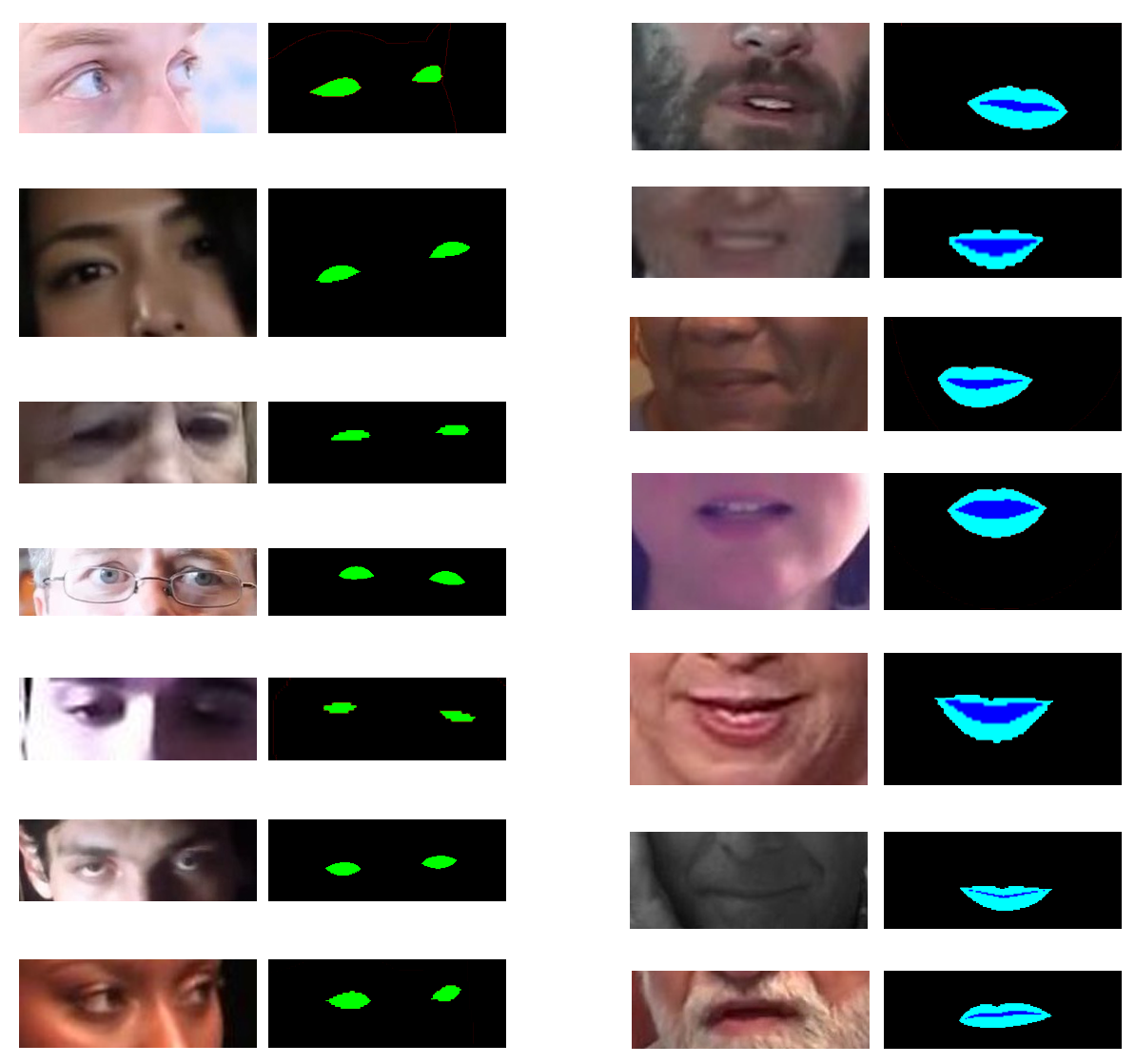}
\caption{Several examples for the eye and mouth sub-datasets. The first two columns are pairs of eyes/masks, while the last two columns are the mouths/masks pairs. The colours green/cyan/blue in masks represents the eyes/outer mouth/inner mouth, respectively. (Best seen in colour)}
\label{fig:4}       
\end{figure}

After all the face masks have been generated, we have organised the dataset to suit our experiments. In particular, we have divided each video into short face sequences of one second (30 frames), and then for each video, we have randomly picked up 10\% of its one-second sequences for our experiments. Since the information of adjacent one-second sequences may heavily overlap with each other, which may cause over-fitting problems, and also consider the training efficiency, we only use 10\% one-second sequences instead of all these short clips. For training/validation/testing, we have randomly selected 619/58/80 one-second sequences, which contains 18570/1740/2400 face images in total, from 93/9/12 videos, and the training/validation/testing sets are subject-independent with each other to guarantee a fair evaluation. This dataset is called '300VW-Mask' dataset, and it is the dataset which we used to train the primary model and to evaluate the performance of final predictions.

For the training of these two additional models focusing on eye and mouth regions, we have further generated two sub-datasets from the afore-mentioned 300VW-Mask dataset. Specifically, we have cropped eye and mouth regions out of the 300VW-Mask dataset to form these sub-datasets. For the purpose of robustness, random noises are added during the cropping process, and we have fixed the locations of cropping box for every 5 consecutive frames so that the temporal information within these frames could be better extracted by the ConvLSTM-FCN models. Fig. \ref{fig:4} has plotted some examples of these two sub-datasets.

\subsection{Experimental Framework}
\label{sec:EXP_Frameworks}
\paragraph{Evaluation Metric}
As mentioned in Section \ref{sec:MT_Loss}, mean Intersection over Union (mIoU) is used as the evaluation metric in the field of semantic segmentation, since mIoU is less sensitive to imbalanced data. Note that we ignored the IoU of background pixels in our mIoU calculation to focus the metric on the face mask pixels.  

\paragraph{Baseline Approaches} 
For the baseline approaches, we have compared the performances of the following methods on the 300VW-Mask dataset: 1. The traditional 68-point facial landmark tracking model \citep{kazemi2014one}. 2. The deeplab-V2 model \citep{chen2016deeplab},  3. The VGG-16 Version of FCN \citep{simonyan2014very,long2015fully}, 4. The ResNet-50 Version FCN \citep{he2016deep,long2015fully}, and 5. The ResNet-50 Version FCN + a simple temporal smoothing strategy.

For the facial landmark tracking model, we have used the 68-landmark model released by DLib library \citep{dlib09}. This model has adopted the face alignment algorithm in the work of \citet{kazemi2014one}, and have been trained on the iBUG 300-W face landmark dataset \citep{sagonas2016300}. We have implemented a 68-landmark face tracker with this alignment model using the methods described in \citet{asthana2014incremental}. This face tracker is run on all the testing set sequences, and the 68 output landmark points are then converted into face masks to calculate the mIoU performance, using the same conversion method as we used to generate face mask labels for the 300VW dataset. 

Deeplab-V2 model is one of the most popular deep models in still image segmentation, and we have also evaluated the performance this model as one of the baseline methods. We have adopted the source code implementation released by Deeplab, and we have selected the model based on VGG-16 architecture. 

The performances of FCN models are more relevant as our ConvLSTM-FCN model is based on the FCN architectures. Therefore, we have evaluated two different FCN models: 1. the VGG-16 version FCN, This model is cast from the VGG-16 architecture. 2. the ResNet-50 version FCN. This is the baseline FCN model that we adopted to convert into ConvLSTM-FCN. Section \ref{sec:MT_Model} described details about this FCN model and its conversion into ConvLSTM-FCN model.  

Besides, we have also applied a simple temporal smoothing strategy to the predictions (after Softmax) of the baseline ResNet-50 FCN in order to compare with the temporal smoothing effects introduced by our ConvLSTM-FCN model. This temporal smoothing technique has a time window of five frames, which is the same size with the time window of our ConvLSTM-FCN model, and the weights for each frame in the time window are subject to a Gaussian distribution centred around the current frame with a standard deviation ($\sigma$) of 0.6.

\paragraph{Training ConvLSTM-FCN Models}
Our ConvLSTM-FCN model, as mentioned in Section \ref{sec:MT_Model}, is converted from the baseline FCN model by replacing the classification layer with ConvLSTM layers. Therefore, to simplify the training process, we first trained a baseline FCN model with all the training images without considering the temporal information. And then we converted this learned FCN model into ConvLSTM-FCN, keeping all the weights except the newly-added ConvLSTM layer, and then retrained it with data of video sequences, where the temporal correlations were learned and extracted. 

In particular, the 300VW-Mask dataset was used to train the primary model. A baseline FCN was first trained on this dataset using cross-entropy loss, and this learned model was used as a reasonable starting point for the training of the primary ConvLSTM-FCN model. For the primary model, we have explored how the applications of ConvLSTM layer and Segmentation Loss could enhance the model's performances by freezing all other layers except the ConvLSTM layer. After this exploration, we used Segmentation Loss to train the primary model by applying different learning rates on the ConvLSTM layer and other layers. Therefore, the training of the ConvLSTM-FCN model was performed as a two-step procedure: first, a baseline FCN model was trained with cross-entropy loss, then this learned model was converted to a ConvLSTM-FCN model to be trained with Segmentation Loss. 

We have utilised similar training strategies for the additional eye and mouth models. Namely, we also first trained a baseline-FCN model focusing on the still eye and mouth images, and then a ConvLSTM-FCN with pre-trained weights was trained to capture the temporal dependencies. 

\paragraph{Implementations}
We built and trained our model under the deep-learning frameworks of Keras \citep{chollet2015keras} and TensorFlow \citep{tensorflow2015-whitepaper}. The models are trained on a desktop with a 1080Ti graphics card and also on a cluster with 10 TITAN X graphics cards. It took around three days to obtain the final primary and additional models. 

For the model training, we have adopted the Adam optimiser \citep{kingma2014adam}, and model's weights were saved and evaluated on the validation set after each epoch. The model with highest validation mIoU was then considered as the best one and was further evaluated on the testing set. All images were resized to 320 by 320 before they were fed into the model. For evaluations on the testing set, model's output heat-map, whose size is also 320 by 320 pixels, was first resized back to the image's original resolution, so that the IoU was calculated at this original scale. 

The baseline model FCN was trained for a total of 80 epochs with batch size 16, learning rate 0.001 with linear decays and cross-entropy loss. The weights of the trained FCN model were then used as the starting point for the ConvLSTM-FCN model, which were trained for another 60 epochs using Segmentation Loss. The learning rate for ConvLSTM-FCN model was layer-based, which was 0.001 for the ConvLSTM layers and  $0.001\gamma$ for other layers, where $\gamma$ is a decaying factor for learning rate. The intuition is to train the newly-added ConvLSTM layer at larger steps while fine-tuning these learned layers with comparatively smaller learning rate. 

For the ConvLSTM layer, the time dimension T was set to be 5, i.e. the ConvLSTM layer deals with short sequences of 5 frames. Therefore, input data of one batch should contain $N\times 5$ images, where N is an integer. In our experiments, we have set N=2, i.e. we have two 5-frame sequences in each batch. 

In the step of integrating the predictions from primary and additional models, we first used the face masks from the primary model to approximately localise the eye and mouth regions for all frames, and then we fixed the cropping box of such regions for each 5-frame sequence so that the additional model could work smoothly to extract temporal information from these short sequences. 

For each experiment, to verify its improvements on the baseline method, we also calculated whether it is statistically significant with the baseline FCN model. Particularly, we split the testing set, which contains 80 one-second sequences, into 10 groups, and calculated the P value of these 10 groups between the current experiment and the baseline model. If the P value is smaller than 0.05, then we consider this experimental result to be statistically significant from that of baseline approach. 

\subsection{Results}
\label{sec:EXP_Results}

\paragraph{Baseline Approaches}
Table \ref{tab:2} shows the performances of the five baseline approaches described in Section \ref{sec:EXP_Frameworks}. The mIoU listed in the table is the average IoU of all classes except the background. It could be seen that although the face tracker approach has achieved the highest mIoU, its prediction for facial skin is worst than other deep methods. The performances of Deeplab-V2 model generally surpasses that of two FCN models, mainly on the eye and inner mouth predictions. These two FCN models achieved similar performances, giving the best facial skin predictions. All these deep models were trained with cross-entropy loss, and the trained model of FCN-ResNet50, which obtains 54.50\% mIoU, would be converted into ConvLSTM-FCN model for further explorations. This trained model of FCN-ResNet50 will be simply called 'baseline-FCN' for convenience. As for the application of temporal smoothing technique, it does not actually improve the performance of the baseline-FCN, and this indicates that the simple temporal smoothing could not properly capture the temporal structure within the consecutive frames.

\paragraph{Exploring ConvLSTM layer}
\begin{table}[t!]
\caption{The IoU performances of baseline approaches. Mean IoU does not take the IoU of background class into consideration. 'FS','OMT','IMT' and 'BG' in the first row is short for facial skin, outer mouth, inner mouth and background. The temporal smoothing approach takes a five-frame time window, and the weights are subject to a Gaussian distribution centred around the current frame ($\sigma=0.6$).}
\label{tab:2} 
\begin{tabularx}{\columnwidth}{ b  k k  k  k k  k}
\hline
Methods & mIoU & FS & Eyes & OMT & IMT & BG \\ \hline
Face Tracker & 60.09 & 88.77 & 50.01 & 61.04 & 40.56 & 97.71 \\ \hline
Deeplab-V2 & 58.66 & 90.55 & 50.19 & 58.58 & 35.31 & 94.38 \\ \hline
FCN-VGG16 & 55.71 & 91.12 & 44.18 & 58.60 & 28.95 & 94.87 \\ \hline
FCN-ResNet50 (Baseline-FCN)& 54.50 & 91.13 & 45.54 & 57.14 & 24.20 & 94.98 \\ \hline
FCN-ResNet50 + Temporal Smoothing& 54.21	& 91.17	& 44.84	& 57.10	 & 23.74 & 95.00 \\ \hline
\end{tabularx}
\end{table}

\begin{table}[t!]
\caption{The IoU performances of Adam and RMSprop optimiser when all layers are frozen except the ConvLSTM layer. Mean IoU does not include the IoU of background class. '$\dagger$' denotes that the difference with the baseline-FCN is statistically significant.}
\label{tab:3} 
\begin{tabularx}{\columnwidth}{ b  k  k   k   k  k  k}
\hline
Optimiser & mIoU & FS & Eyes & OMT & IMT & BG \\ \hline
Adam & 55.53$^\dagger$ & 91.07 & 45.70 & 57.58 & 27.78 & 94.85 \\ \hline
RMSprop & 54.93 & 91.31 & 46.52 & 58.70 & 23.20 & 94.98\\ \hline
\end{tabularx}
\end{table}

\begin{table}[t!]
\caption{The mean improvement over baseline-FCN in time dimension. Larger value indicates greater improvements over the baseline-FCN. 'T1' to 'T5' represents the first frame to the last (fifth) frame for video sequences of five frames.}
\label{tab:4} 
\begin{tabularx}{\columnwidth}{s  s  s  s  s s}
\hline
Optimiser & T1 & T2 & T3 & T4 & T5  \\ \hline
Adam & 0.113 & 0.964  & 0.995 & 0.950 & 0.981  \\ \hline
RMSprop & 0.019  & 0.174 & 0.275 & 0.598 & 0.657 \\ \hline
\end{tabularx}
\end{table}

As mentioned in Section \ref{sec:EXP_Frameworks}, we have made some explorations in order to see if the ConvLSTM layer could actually improve the performance by using temporal information. For simplicity, after the baseline-FCN model was converted into ConvLSTM-FCN, we have frozen all other layers and only trained the newly-added ConvLSTM layer with  cross-entropy loss. We have also tried two optimisers: Adam \citep{kingma2014adam} and RMSprop \citep{hinton2012rmsprop}. The results are shown in Table \ref{tab:3}. It could be seen that the Adam and RMSprop optimisers both improve the mIoU slightly. For further validation, we have also computed their improvement over the baseline-FCN on the time dimension T, which is 5 in our ConvLSTM-FCN model. It could be seen in Table \ref{tab:4} that, for all 5-frame sequences, the improvements on the last four frames is generally higher than that of the first frame, which indicates the ConvLSTM layer can actually extract temporal information from video sequences to improve segmentation accuracy. Besides, it is also interesting to observe that the temporal smoothing effects are more obvious in the RMSprop experiment, with incremental improvements as time dimension increases. 

Therefore, by these exploration experiments, we have verified that ConvLSTM could actually produce temporal smoothing effects for face mask extraction in video sequences. We have also selected Adam as the optimiser for following experiments.

\paragraph{Segmentation Loss}
\begin{table}[t!]
\caption{The performances of cross-entropy loss, IoU Loss and the proposed Segmentation Loss (SegLoss). The IoU Loss (defined in Eq. \ref{eq:9}) is  the multiple-class version extended from the binary work by \citet{rahman2016optimizing}. For Segmentation Loss, we have tested different forms of loss calculation function, i.e. $L_p(x_i)$ and $L_n(x_i)$ in Eq. \ref{eq:6}. All layers except the newly-added ConvLSTM layer is frozen. Mean IoU does not include the IoU of background class. '$\dagger$' denotes that the difference with the baseline-FCN is statistically significant.}
\label{tab:5} 
\begin{tabularx}{\columnwidth}{ b  k  k   k   k  k  k}
\hline
Loss Definitions & mIoU & FS & Eyes & OMT & IMT & BG \\ \hline
Cross-Entropy Loss & 55.53$^\dagger$ & 91.07 & 45.70 & 57.58 & 27.78 & 94.85 \\ \hline
IoU Loss (Eq. \ref{eq:9}) & 57.22$^\dagger$ & 91.06 & 49.02 & 57.86 & 30.95 & 94.88\\ \hline
Seg Loss (Eq. \ref{eq:7}, g=1) & 58.10$^\dagger$ & 90.90 & 51.96 & 59.32 & 30.20 & 94.82 \\ \hline
Seg Loss (Eq. \ref{eq:8}, g=0.1) & 58.39$^\dagger$ & 90.56 & 51.45 & 57.43 & 34.11 & 94.76\\ \hline
Seg Loss (Eq. \ref{eq:8}, g=0) & 59.04$^\dagger$ & 90.80 & 51.61 & 57.27 & 36.46 & 94.91\\ \hline
\end{tabularx}
\end{table}
We have also conducted experiments to explore to what extend the proposed Segmentation Loss can lead to better a performance for the ConvLSTM-FCN model. As explained in Section \ref{sec:MT_Loss}, the loss calculation function for positive and negative samples, which is $L_p(x_i)$ and $L_n(x_i)$ in Eq. \ref{eq:6}, could have various potential definitions, and we have provided two forms of them in Eq. \ref{eq:7} and Eq. \ref{eq:8}. For the simplicity of the experiments, we have employed the same strategy as in the experiments of exploring ConvLSTM layer, i.e. after casting the baseline-FCN into ConvLSTM-FCN model, all other layers are frozen and the only trainable layer is the newly-added ConvLSTM layer. Then we used Segmentation Loss to train this partially-frozen ConvLSTM-FCN model. For comparison, we have also evaluated the performances of the cross-entropy loss and the multiple-class IoU loss defined in Eq. \ref{eq:9}. Table \ref{tab:5} summarises the results, and it could be seen that both IoU Loss and Segmentation Loss achieve higher mIoUs than the cross-entropy loss, however, our Segmentation Loss shows the best performances in all the three losses, no matter which kind of loss calculation function is used. This demonstrates the effectiveness of the proposed Segmentation Loss in terms of optimising the ConvLSTM-FCN model. In addition, the loss function $L_p(x_i)$ and $L_n(x_i)$ defined Eq. \ref{eq:8} have shown the best mIoU performance when $g$ is 0, therefore, we have selected the form in Eq. \ref{eq:8} ($g=0$) for Segmentation Loss in  the following experiments.

\paragraph{Training Primary and Zoomed-in Models}
As mentioned in Section \ref{sec:EXP_Frameworks}, We have applied similar strategies to train the primary and additional models. For the primary model, after the baseline-FCN was transformed into ConvLSTM-FCN, we have set different learning rates for different layers, which is 0.001 for ConvLSTM layer and $0.001 \gamma$ $(\gamma \in (0,1))$ for other layers, since we would like the newly-added ConvLSTM layer to learn faster than other already-trained layers. The Segmentation Loss with $L_p(x_i)$ and $L_n(x_i)$ defined in Eq. \ref{eq:8} (g=0) is used to train the primary ConvLSTM-FCN model. Table \ref{tab:6} has demonstrated the performances of the primary model with different $\gamma$ values. It could be seen that different $\gamma$ values could slightly affect the performances, while training ConvLSTM-FCN model with different internal learning rates could generally achieve better mIoUs than just freezing all layers except ConvLSTM layer.  

\begin{table}[t!]
\caption{The IoU performances of the primary model with different $\gamma$ values. The ConvLSTM layer is trained with learning rate 0.001, while the learning rate of other layers are set to be 0.001$\gamma$. Mean IoU does not include the IoU of background class. '$\dagger$' denotes that the difference with the baseline-FCN is statistically significant.}
\label{tab:6} 
\begin{tabularx}{\columnwidth}{ s  s  s   s   s  s  s}
\hline
$\gamma$ & mIoU & FS & Eyes & OMT & IMT & BG \\ \hline
0.01 & 60.35$^\dagger$ & 89.83 & 56.50 & 59.61 & 35.45 & 93.79 \\ \hline
0.02 & 60.96$^\dagger$ & 89.85 & 57.75 & 60.02 & 36.23 & 93.72\\ \hline
0.05 & 60.04$^\dagger$ & 90.46 & 54.89 & 58.98 & 35.86 & 94.36\\ \hline
0.1 & 60.07$^\dagger$ & 90.51 & 54.73 & 59.74 & 35.29 & 94.42\\ \hline
\end{tabularx}
\end{table}

\begin{table}[t!]
\centering
\caption{The IoU performances of the additional eye model on the sub-dataset of eyes. The ConvLSTM layer is trained with learning rate 0.001, while the learning rate of other layers are set to be 0.001$\gamma$. '$\dagger$' denotes that the difference with the baseline-FCN is statistically significant.}
\label{tab:7} 
\begin{tabularx}{\columnwidth}{bss}
\hline
Model & Eyes & BG \\ \hline
Baseline-FCN & 54.29 &    98.23 \\ \hline
ConvLSTM-FCN ($\gamma=0.01$) & 56.58 &    98.25  \\ \hline
ConvLSTM-FCN ($\gamma=0.02$) & 
59.01$^\dagger$ & 98.14  \\ \hline
ConvLSTM-FCN ($\gamma=0.05$) & 
57.51$^\dagger$ & 98.24 \\ \hline
ConvLSTM-FCN ($\gamma=0.1$) & 
51.82$^\dagger$ & 97.88 \\ \hline
\end{tabularx}
\end{table}

\begin{table}[t!]
\centering
\caption{The IoU performances of the additional mouth model on the mouth sub-dataset. The ConvLSTM layer is trained with learning rate 0.001, while the learning rate of other layers are set to be 0.001$\gamma$. Mean IoU does not include the IoU of background class. '$\dagger$' denotes that the difference with the baseline-FCN is statistically significant.}
\label{tab:8} 
\begin{tabularx}{\columnwidth}{bkkkk}
\hline
Model & mIoU & OMT & IMT & BG \\ \hline
Baseline-FCN & 49.77 &    60.30 & 39.24 & 97.23 \\ \hline
ConvLSTM-FCN ($\gamma=0.01$)  & 52.08$^\dagger$ & 62.06 & 42.10 & 97.21 \\ \hline
ConvLSTM-FCN ($\gamma=0.02$) & 52.17$^\dagger$ & 62.80 & 41.54 & 97.31 \\ \hline
ConvLSTM-FCN ($\gamma=0.05$) & 51.24$^\dagger$ & 61.01 & 41.48 & 97.15 \\ \hline
ConvLSTM-FCN ($\gamma=0.1$) & 52.36$^\dagger$ & 61.20 & 43.52 & 96.86 \\ \hline
\end{tabularx}
\end{table}

Similarly, for the additional models on eye and mouth regions, we first used cross-entropy loss to train two baseline-FCN models on the eye and mouth sub-datasets, respectively, and these baseline-models are then converted into ConvLSTM-FCN models, which are also trained with different internal learning rates, as in the primary model's training. Table \ref{tab:7} and Table \ref{tab:8} show the performances of baseline-FCN and ConvLSTM-FCN with different $\gamma$ values. It can be seen from the results that ConvLSTM-FCN model with Segmentation Loss could generally improve the performance of the baseline-FCN model, and the additional model focusing on certain face region could achieve better segmentation accuracy on that region than that of the primary model. 

\begin{figure*}[ht!]
\captionsetup{justification=centering}
\includegraphics[width=\textwidth]{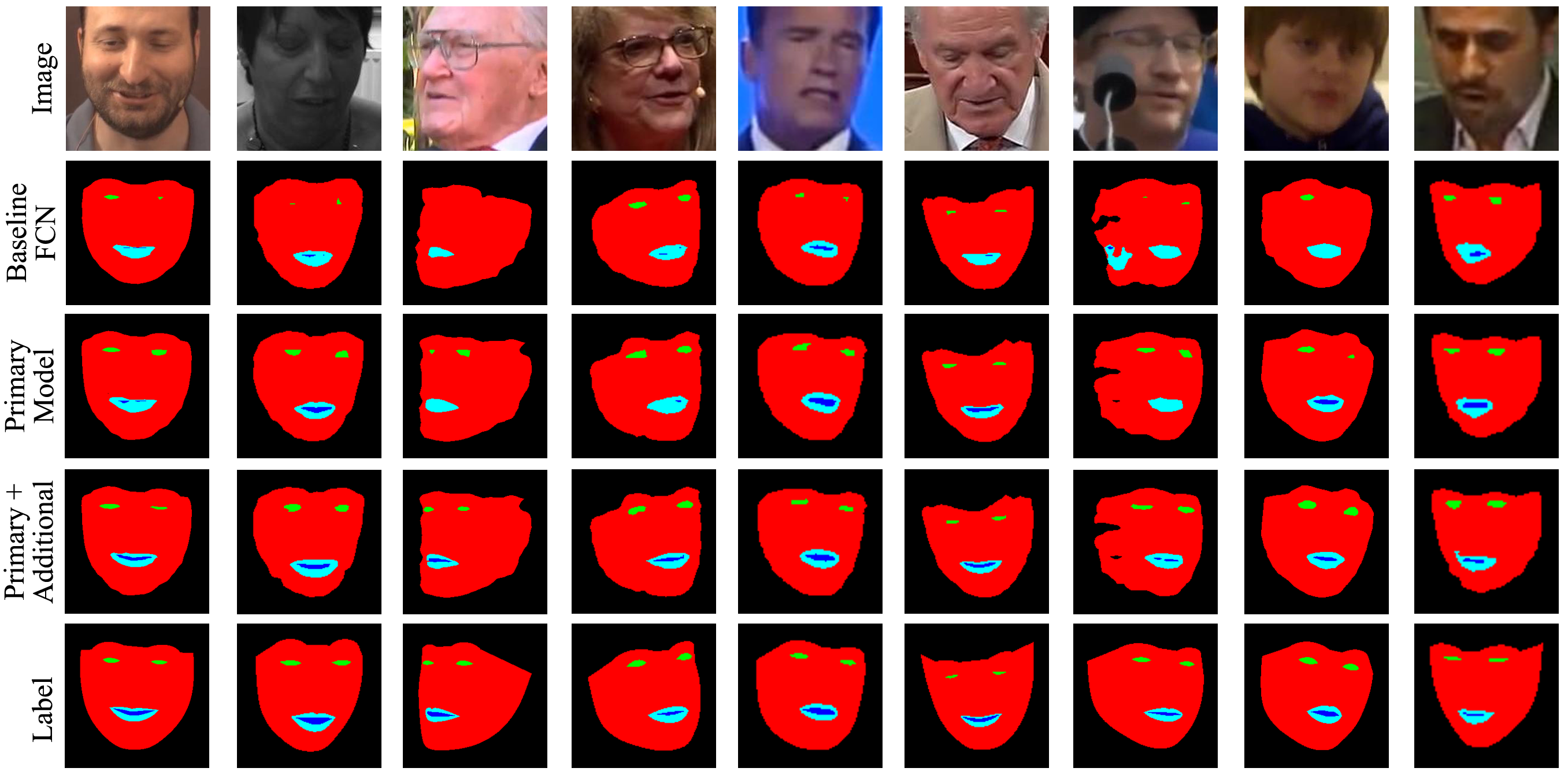}
\caption{Several face masks extracted with baseline-FCN, the primary model and the integration of primary and additional models. The colours red/green/cyan/blue in face masks stands for facial skin/eye/outer mouth/inner mouth, respectively. (Best seen in colour)}
\label{fig:5}       
\end{figure*}

\begin{table}[t!]
\caption{The IoU performances of different key techniques on improving the baseline-FCN models. Mean IoU does not include the IoU of background class. '$\dagger$' denotes that the difference with the baseline-FCN is statistically significant. The temporal smoothing approach takes a five-frame time window, and the weights are subject to a Gaussian distribution centred around the current frame ($\sigma=0.6$). The IoU Loss (defined in Eq. \ref{eq:9}) is  the multiple-class version extended from the binary work by \citet{rahman2016optimizing}. The primary model is the ConvLSTM-FCN model trained with 300VW-Mask dataset ($\gamma = 0.05$), and the two additional models are the  ConvLSTM-FCN model trained on two sub-datasets on eye and mouths ($\gamma = 0.02$).}
\label{tab:9} 
\begin{tabularx}{\columnwidth}{ b  k  k   k   k  k  k}
\hline
Techqniques & mIoU & FS & Eyes & OMT & IMT & BG \\ \hline
FCN-ResNet50 + cross-entropy& 54.50 & 91.13 & 45.54 & 57.14 & 24.20 & 94.98 \\ \hline
FCN-ResNet50 + cross-entropy + Temporal Smoothing& 54.21	& 91.17	& 44.84	& 57.10	 & 23.74 & 95.00 \\ \hline
ConvLSTM-FCN (Freezing Other Layers) + cross-entropy & 55.53$^\dagger$ & 91.07 & 45.70 & 57.58 & 27.78 & 94.85 \\ \hline
ConvLSTM-FCN (Freezing Other Layers) + IoU Loss & 57.22$^\dagger$ & 91.06 & 49.02 & 57.86 & 30.95 & 94.88 \\ \hline
ConvLSTM-FCN (Freezing Other Layers) + Segmentation Loss & 59.04$^\dagger$ & 90.80 & 51.61 & 57.27 & 36.46 & 94.91\\ \hline
Primary Model + Segmentation Loss & 60.04$^\dagger$ & 90.46 & 54.89 & 58.98 & 35.86 & 94.36\\ \hline
Primary Model + Two Additional Models + Segmentation Loss & \textbf{63.76$^\dagger$} & 90.58 & 57.89 & 62.78 & 43.79 & 94.36\\ \hline
\end{tabularx}
\end{table}

\paragraph{Integrating Predictions} 
As described in Section \ref{sec:MT_PAM} and Section \ref{sec:EXP_Frameworks}, the final predictions are obtained by integrating the face masks of the primary model, which provides localisations of eye and mouth regions, with the corresponding outputs of two additional models on the eye and mouth regions. These additional models focus on particular facial parts, such as eyes, outer and inner mouths, therefore they could produce more accurate segmentation results for these regions. 

For the final predictions, we have used the primary model which are trained with $\gamma=0.05$, and the ConvLSTM models trained with $\gamma=0.02$ for eye and mouth additional models (the performances of these models could be found in Table \ref{tab:6}, Table \ref{tab:7} and Table \ref{tab:8}).

The integration results could be found in Table \ref{tab:9}, and this table also summarises the key improvements on the baseline-FCN model with different techniques. It can be seen from the table that the application of a simple temporal smoothing technique could not actually improve the performances of the baseline-FCN model, as it cannot appropriately capture the inherent temporal structure within video sequence. Our ConvLSTM-FCN model, however, shows an 1.03\% abosolute improvement over the baseline-FCN model, even when all other layers except the ConvLSTM layer are froze and are trained with cross-entropy loss, which validates the introduced temporal smoothing effective from ConvLSTM-FCN model. Besides, combining primary model and additional models leads to a mIoU performance of 63.76\%, which shows a 16.99\% relative improvement on the baseline-FCN approach. When compared with these baseline approaches in Table \ref{tab:2}, our proposed method still shows higher segmentation accuracies, even with the face tracker, which is the best-performing baseline approach. 

\subsection{Discussion}
\label{sec:EXP_Discussion}

\begin{figure*}[ht!]
\captionsetup{justification=centering}
\includegraphics[width=\textwidth]{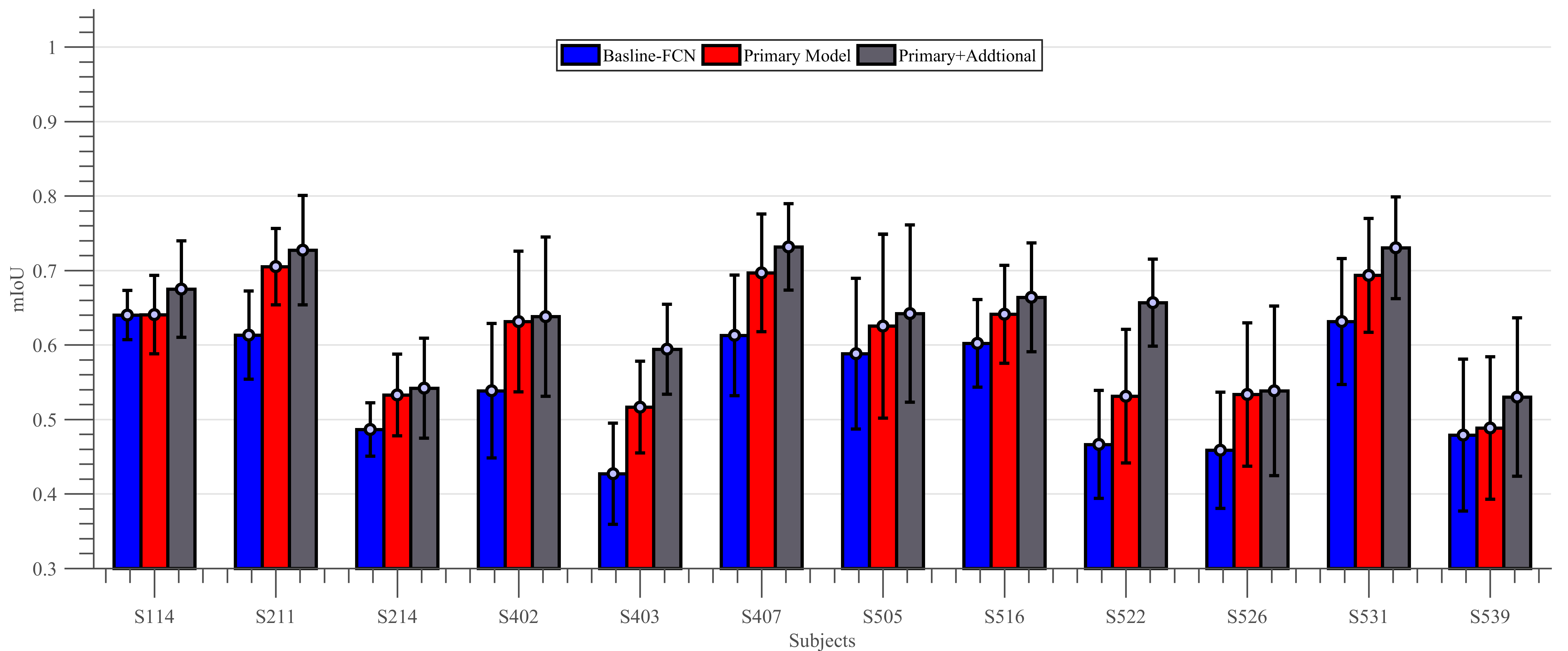}
\caption{Mean IoU and standard deviation  over all frames of each subject. Mean IoU does not include the IoU of background class. Blue stands for the performances of baseline-FCN, red for the primary model and gray for the integration of primary and additional models. (Best seen in colour)}
\label{fig:6}       
\end{figure*}

In the task of face mask extraction, the temporal dimension carries important information which could be utilised to improve segmentation accuracies, especially when the information provided by current frame is not sufficient to allow reliable face mask extraction. This temporal-smoothing effect is what we would like to achieve with our ConvLSTM-FCN model. 

In the case when normal FCN models encounter challenging segmentation tasks, the introduced ConvLSTM-FCN should be able to achieve better performances by exploiting information from both temporal and spatial domains. Fig. \ref{fig:5} plots some typical examples of such situations. As shown in the figure, the baseline-FCN model, which only learns the spatial relationships, have difficulties in segmenting face images with low qualities, occlusions, poor illuminations, etc. As a result, baseline-FCN could not effectively segment those smaller facial regions such as eyes and inner mouth under challenging scenarios. However, with the help of ConvLSTM-FCN model, the extracted face masks are more robust and realistic, especially for the smaller facial regions like eyes and inner mouth. The introduction of the zoomed-in model has further improved the segmentation results, which again verify the temporal-smoothing effects introduced by ConvLSTM-FCN. 

Fig. \ref{fig:6} shows the mean IoU performances and standard deviation over all frames of each subject for the baseline-FCN, primary model and the integration of primary \& additional models. The test set contains 80 one-second sequences coming from 12 videos, while these 12 videos are subject-independent with each other. It could be observed that the primary model or primary + additional have led to better performances than baseline-FCN on all the subjects. Besides, we could also see that the performances over different test subjects are generally similar, despite some fluctuations brought by the video variations.

\section{Conclusion}
\label{sec:Conclusion}
In this paper, we have presented a novel ConvLSTM-FCN model for the task of face mask extraction in video sequences. We have illustrated how to convert a baseline-FCN model into ConvLSTM-FCN model, which can learn from both temporal and spatial domains. A new loss function named 'Segmentation Loss' has also been proposed for training the ConvLSTM-FCN model. Last but not least, we also introduced the engineering trick of supplementing the primary model with two zoomed-in models focusing on eyes and moth. With all these are combined, we have successfully improved the performances of baseline-FCN on 300VW-Mask dataset from 54.50\% to 63.76\%, making a 16.99\% relative improvement. The analysis of the experimental results has verified the temporal-smoothing effects brought by the ConvLSTM-FCN model.

\bibliographystyle{spbasic}      
\bibliography{ref.bib}

\end{document}